\documentclass[letterpaper, 10 pt, conference]{ieeeconf}  

\usepackage{booktabs}
\usepackage{amsmath}
\usepackage{amssymb} 
\usepackage{color}
\usepackage{xcolor}
\usepackage{epsfig}
\usepackage{graphicx}
\usepackage{soul}

\newcommand{\etal}{\textit{et al}. }
\newcommand{\Xk}{\mathbf{X}_k}
\newcommand{\Fk}{\mathbf{F}_k}
\newcommand{\Ek}{\mathbf{E}_k}
\newcommand{\Ik}{\mathbf{I}_k}
\newcommand{\Pc}{\mathbf{P}}
\newcommand{\Ep}{\mathbf{E}_P}
\newcommand{\Pk}{\mathbf{P}_k}
\newcommand{\Dk}{\mathbf{D}_k}

\IEEEoverridecommandlockouts                              

\title{\LARGE \bf
LAPTNet-FPN: Multi-scale LiDAR-aided Projective Transform \\ Network for Real Time Semantic Grid Prediction
}

\author{Manuel Diaz-Zapata$^{1,2}$, David Sierra-Gonzalez$^{2}$, \"{O}zg\"{u}r Erkent$^{2,3}$, Christian Laugier$^{2}$, Jilles Dibangoye$^{1,2}$   
\thanks{This work was partially supported by EU project CPS4EU and Toyota Motor Europe.}
\thanks{ 
$^{1}$CHROMA team, Univ Lyon, Inria, INSA Lyon, CITI Lab, France. $^{2}$CHROMA team, Univ. Grenoble Alpes, Inria, Grenoble, France.  $^{3}$Hacettepe University, Ankara, Turkey.
Correspondence: {\tt\small manuel.diaz-zapata@inria.fr}}%
}

\begin{document}

\maketitle
\thispagestyle{empty}
\pagestyle{empty}

\begin{abstract}
Semantic grids can be useful representations of the scene around an autonomous system. By having information about the layout of the space around itself, a robot can leverage this type of representation for crucial tasks such as navigation or tracking. By fusing information from multiple sensors, robustness can be increased and the computational load for the task can be lowered, achieving real time performance. Our multi-scale LiDAR-Aided Perspective Transform network uses information available in point clouds to guide the projection of image features to a top-view representation, resulting in a relative improvement in the state of the art for semantic grid generation for  human (+8.67\%) and movable object (+49.07\%) classes in the nuScenes dataset, as well as achieving results close to the state of the art for the vehicle, drivable area and walkway classes, while performing inference at 25 FPS. 

\end{abstract}

\section{INTRODUCTION}

    
The top-view representation of a scene, also known as bird's eye view (BEV), can be an extremely useful tool for autonomous systems that are limited to the ground plane, as it is in the case for wheeled or legged robots. Having a representation of the scene in the BEV can be beneficial to downstream tasks such as tracking, prediction and navigation. Compared to the camera point of view, the size of the objects in the BEV does not vary depending on their distance to the robot. Semantic grids are one type of BEV representation that allow a dense representation of the space around the robot. In a semantic grid, each cell contains a semantic label corresponding to the class of the object or area occupying it.

The interest over semantic grid prediction has increased in recent years. Existing approaches in the literature can be grouped along three lines: camera-based, LiDAR-based and sensor fusion based methods. For camera-based methods, the main challenge is how the projection to the BEV is performed from the image plane. Some methods rely on MLPs to find the relationships between camera plane and BEV \cite{roddick-pon, vpn, ved} . Others perform categorical depth prediction to do this projection \cite{philion-lss, fiery}, whereas Xie \etal assume an uniform depth distribution along the camera ray \cite{m2bev}. More recently, the use of attention mechanisms has also been proposed as a solution to this challenge \cite{imgs2map, cross-view-transformers, bevformer}. For LiDAR-based approaches, some follow a bayesian approach \cite{cmcdot}, whereas others leverage geometric relationships in the point cloud as well as occupancy features \cite{pillarsegnet}. Sensor fusion approaches leverage information from different modalities such as camera and LiDAR by processing each one with separate encoders and then fusing them via heuristics \cite{hendy2020fishing}, convolutions \cite{erkent2019end,bevfusion} or self-attention mechanisms \cite{transfusegrid}.

In the proposed LAPTNet-FPN, we use the geometric information about the scene coming from a LiDAR sensor to guide the projection of camera features. By doing this projection at multiple image scales given by a convolutional backbone, we are able to associate BEV cells to pixels in the images and do semantic grid prediction in real time.

\section{RELATED WORK}
The utility of the BEV representation space can be seen in its increased use for different tasks such as 3D bounding box detection and map segmentation in the robotics community during recent years.

\subsection{BEV for 3D bounding box detection}
 Regarding camera-based approaches, the Orthographic Feature Transform (OFT) creates a BEV representation from an intermediate voxel representation using camera information \cite{oft}. CaDDN and BEVDet predict a categorical depth distribution to project image features from the camera plane to the BEV \cite{caddn, bevdet}. PseudoLidar predicts a point cloud using camera images, which is then projected to the BEV where height is encoded in the channel dimension \cite{wang2019pseudo} . 
 
 Aside from camera-only methods, MV3D  uses the BEV representation space to encode the LiDAR point cloud information which is then fused with camera features \cite{mv3d}. In contrast, PointAugmenting fetches image features to enrich point cloud data which are processed by a 3D backbone and then flattened to the BEV \cite{wang2021pointaugmenting}.

\subsection{BEV semantic grid prediction}
Although the BEV has been successfully used for the task of 3D bounding box estimation, the current trend for the use of this representation space is to generate semantic grids (also known as semantic maps). For this task, the current literature can be classified into three main types depending on the input data modalities: camera-based, LiDAR-based or sensor fusion based grid generation.

\subsubsection{Camera-based grid generation}
Generation of semantic grids solely from camera information is a challenging task. This is mainly due to the difficulty of correlating information between the camera plane and the BEV given the pinhole camera projection model \cite{cv-book} and the lack of geometric information about the scene. Some methods such as Lift-Splat-Shoot (LSS), predict a probability distribution for a set of discrete depth values in order to project the image features to 3D space; these features are then sum-pooled to generate the BEV map that is used for semantic grid generation \cite{philion-lss}. FIERY follows the same projection method as LSS and accumulates these features in the BEV along a temporal dimension to create the semantic grids \cite{fiery}. Compared to LSS, M2BEV assumes a uniform depth distribution along the camera ray to project the image features to a voxel space that is then reduced to a BEV using convolutions \cite{m2bev}. Others, like the Pyramid Occupancy Network, the View Parsing Network (VPN) and the Variational Encoder-Decoder Network use fully-connected networks to 'shrink' the 2D camera information into 1-dimensional feature vectors that are then sampled to generate the 2D BEV  \cite{roddick-pon,vpn,ved}. 

Attention mechanisms have also been proposed as a method to correlate the information from the camera plane to the BEV plane. Saha \etal generate the BEV from image features by projecting vertical scanlines in the image plane to polar rays in the top-view using inter-plane attention and polar ray self-attention mechanisms \cite{imgs2map}. Zhou and Krähenbühl propose a camera-aware, cross-view attention mechanism to learn the mapping from each camera image to the BEV using the camera parameters \cite{cross-view-transformers}. BEVFormer proposes the use of deformable attention \cite{deformable-attn} to generate the BEV feature map from features across different image scales and timesteps \cite{bevformer}.

\subsubsection{LiDAR-based grid generation}
Given the 3D nature of the point clouds generated by a LiDAR sensor, the projection step from the input to the 2D BEV representation becomes trivial. Instead, two of the challenges of working with point clouds for the generation of semantic grids are: the data sparsity in point clouds compared to camera images and the lack of color and texture information of the scene. In the case of PillarSegNet, semantic grids are generated by using a pseudoimage \cite{pointpillars} and occupancy feature maps, both in the BEV \cite{pillarsegnet}. 

\subsubsection{Fusion-based grid generation}
Being aware of the drawbacks of using only one or the other sensor modality, some methods propose a sensor fusion approach. Erkent \etal  map the semantic segmentation predictions from camera images to the BEV using a set of intermediary planes, concatenate them to grids generated by an occupancy grid filter \cite{cmcdot}, and fuse them together using convolutions \cite{erkent2019end}. FISHINGNet  projects camera features to the BEV using VPN and predicts a set of BEV semantic maps using separate backbones for each modality (LiDAR, camera and radar) \cite{hendy2020fishing}. The final semantic output is generated using heuristics to fuse the results of each modality. BEVFusion uses LSS to generate a BEV from camera images and VoxelNet \cite{second} for LiDAR \cite{bevfusion}. These features are then concatenated and fused using convolutions to generate the final semantic grid. TransFuseGrid proposes using a series of self-attention operations to fuse BEV feature maps generated with LSS for camera images and PointPillars for LiDAR point clouds \cite{transfusegrid}. Although these works leverage the BEV from diferent sensor modalities, all of them rely on camera-based approaches to perform the projection from the camera image plane to the BEV plane. 

Our approach, the multi-scale LiDAR-aided Projective Transform Network (LAPTNet-FPN), differs from the state of the art by the fact that we leverage LiDAR information to do the projection of the camera information for semantic grid prediction. Compared to PointAugmenting \cite{wang2021pointaugmenting}, we propose to do this step across with multiple image scales instead of a single one. The use of multiple scales allows us to populate more cells in the BEV with the image information improving semantic grid prediction. 

We list our contributions as follows:
\begin{itemize}
    \item A novel method for real time semantic grid prediction using camera and LiDAR information.
    \item An efficient approach for projection of multiple image scales to a bird's eye view representation.
    \item Extensive ablation studies that explore how the performance changes when using multiple image scales, projecting to different grid resolutions, as well as how it is improved by adding a parallel LiDAR-dependent backbone to combine BEV representations.
    \item A study exploring the dependence on the camera and LiDAR modalities, as well as the performance under adverse conditions such as night and rain.
\end{itemize}

\section{LAPTNet-FPN: Multi-scale LiDAR-aided Projective Transform Network}

\begin{figure*}[t]
  \centering
  \includegraphics[width=\linewidth]{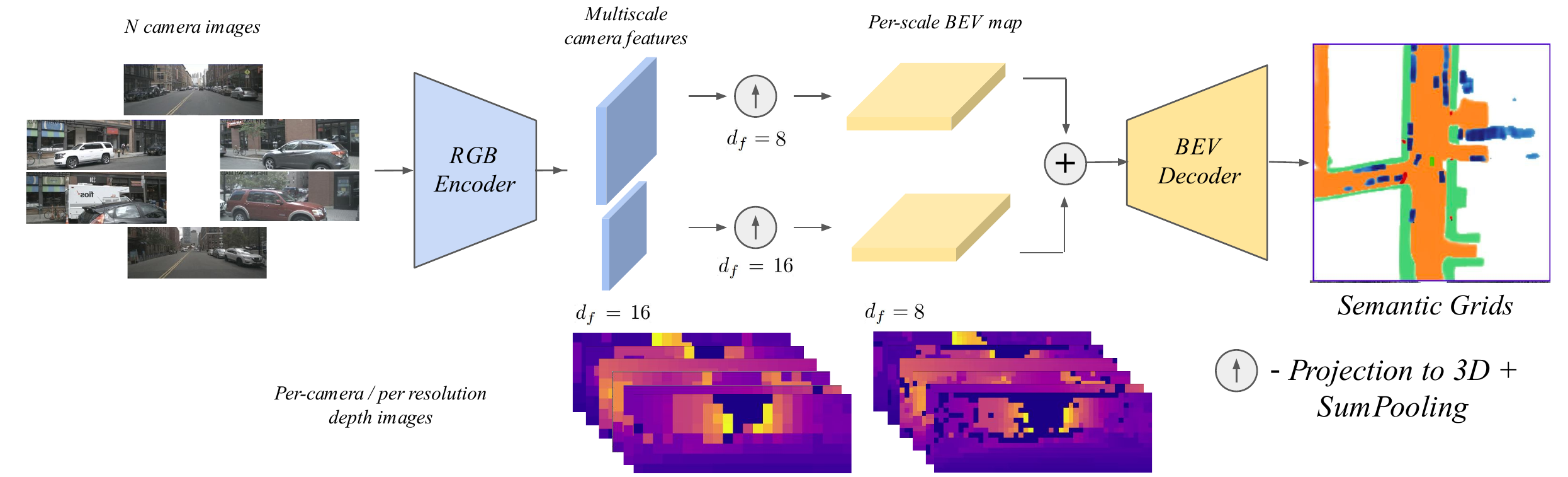}
   \caption{Proposed architecture for the LAPTNet-FPN. We reduce the spatial size of the camera images with a  convolutional RGB backbone taking the last two feature scales with downsampling factors 8 and 16. Depth images are generated by projecting the point cloud to each camera view and applying a minpooling operation to match the spatial size of the image features. Using this information, and a sum pooling operation, the features are projected to a BEV representation that is fed to a convolutional decoder for the generation of the semantic grids. The output shows a superposition of the individually trained classes for a sample. Best viewed with digital zoom.}
  \label{fig:arch}
\end{figure*}

In this section, we explain how the LiDAR-aided Perspective Transform Network (LAPTNet) leverages point cloud information together with camera parameters to perform the projection of information from an arbitrary number of cameras to an unified bird’s eye view for semantic grid prediction. The network will be presented in three main modules, the camera encoder which extracts image features (subsection \ref{subsec:camencode}), the LiDAR-based projection transform (LAPT) that correlates multi-scale camera features to the BEV (subsections \ref{subsec:lapt} and \ref{subsec:laptfpn}) and the BEV decoder that generates the semantic output from these feature grids (subsection \ref{subsec:bevencode}) The architecture of the network can be seen in Fig. \ref{fig:arch}.

\subsection{Problem formulation}

Given a number of $n$ RGB images $ \{\Xk \in \mathbb{R}^{3 \times H \times W}\}_{n} $ of size $(H \times W)$ taken from cameras placed in a vehicle and a point cloud $\Pc \in \mathbb{R}^{3 \times D}$ with $D$ points taken from a LiDAR sensor located on top of it, we want to estimate an occupancy grid for $C$ classes of size $(X \times Y)$ in the BEV $(\mathbf{y} \in \mathbb{R}^{C \times X \times Y})$ that is centered on the coordinate frame of the vehicle. 

Using the available intrinsic $\Ik$ and extrinsic $\Ek$ camera parameters, together with the transformation matrix of the LiDAR sensor $\Ep$, we project the image features from the camera plane to the BEV by leveraging the geometric information available in the point cloud.

\subsection{Image feature extraction}
\label{subsec:camencode}
For our approach, we want to maximize the ratio of correspondences between pixels in $\Xk$ and the points from $\Pc$ that fall into the field of view for camera $k$. Normally, if we perform this association in the raw input space, we would have six times as many pixels as points \cite{nuscenes}. This is assuming that: 1) all LiDAR rays give a return and 2) the points are located in parts of the scene visible by at least one camera, which is not always the case in practice. This implies that an important amount of information about the scene that is captured by the cameras would not have geometric information associated to it for the projection to the BEV.

Knowing this, we use a convolutional backbone to downsample the images to a size that will allow a higher ratio of correspondences, while encoding color and texture information in the channel dimension. For each image, we decrease the input image shape by a factor of $d_f$ to generate a feature map $\{\Fk  \in \mathbb{R}^{N_{f} \times H/d_f \times W/d_f}\}_n$ containing $N_f$ channel-wise features. Any standard image encoder like Resnet \cite{resnet}, FPN \cite{fpn} or EfficientNet \cite{efficientnet} can be used for this step.

\subsection{LiDAR-aided projective transform (LAPT)}
\label{subsec:lapt}
With the camera images now encoded in a reduced spatial size, we turn our attention on how to use the point cloud for the projection of the found features $\Fk$ to the BEV. 

In comparison to the LiDAR sensor, one camera is not able to get a full 360$^{\circ}$ view of the scene. In order to associate the image features to their position in the BEV, we need to project $\Pc$ to each camera's field of view. This corresponds to a perspective transformation of the 3D points, which requires us to know the camera parameters $\Ik,\Ek$ and transformation matrix of the LiDAR $\Ep$. Using a homogeneous coordinate system, we transform the point cloud from the LiDAR reference frame to the vehicle reference frame using the inverse of $\Ep$, and then to the camera reference frame using $\Ek$ as shown in Eq. \ref{eq:transformLidartoCam}. The points in the camera reference frame $\Pk$ are then normalized to 2D homogeneous coordinates in order to be projected to the image plane of $\Xk$ using the camera's intrinsic matrix $\Ik$ as indicated in Eq. \ref{eq:transform3Dto2D}.  
\begin{equation}
\label{eq:transformLidartoCam}
    \begin{pmatrix} \Pk \\ 1 \end{pmatrix} = \Ek \times \Ep^{-1} \times \begin{pmatrix} \Pc \\ 1  \end{pmatrix} = \Ek \times \Ep^{-1} \times \begin{pmatrix} x_{\Pc} \\ y_{\Pc} \\ z_{\Pc} \\ 1  \end{pmatrix}
\end{equation}

\begin{equation}
\label{eq:transform3Dto2D}
    \begin{pmatrix} u_k \\ v_k \\ 1 \end{pmatrix} = \Ik \times \frac{\Pk}{z_{k}}= \Ik \times \begin{pmatrix} x_k/z_{k} \\ y_k/z_{k} \\ z_{k}/z_{k} \end{pmatrix}%
\end{equation}

In the case that two or more points of $\Pk$ are projected to the same pixel, we assign the pixel to the closest one in the $z_k$ axis, following the pinhole camera model \cite{cv-book}. The result after this projection step is a sparse depth image $\Dk$ with the same spatial size of $\Xk$, but only one channel in which we put the depth values $z_k$ available from $\Pk$. Having now some depth correspondences for the pixels in $\Xk$, we downsample $\Dk$ to match the spatial size of our processed feature map $\Fk$ through a minimum pooling operation with a squared kernel size of $d_f$. We use minimum pooling since we need to find the closest depth correspondence for each pixel in $\Fk$. The corresponding distance values $\{\delta_k \in \mathbb{R}^{1\times H/d_f \times W/d_f}\}_n$ are then used to do the projection for the points in $\Fk$ to the robot's reference frame, as shown in Eq. \ref{eq:transform2Dto3D}-\ref{eq:transform_TFs}. Features that are in image coordinates $(u_f, v_f)^T$ in $\Fk$, will be projected to the point $(x_f, y_f, z_f)^T$ in 3D space.

\begin{equation}
\label{eq:transform2Dto3D}
    \begin{pmatrix}
        x_f \\
        y_f \\
        z_f
    \end{pmatrix}_k = \Ik^{-1} \times \delta_k \times \begin{pmatrix}
        u_f \\
        v_f \\
        1
    \end{pmatrix}_k
\end{equation}

\begin{equation}
\label{eq:transform_TFs}
    \begin{pmatrix}
        x_f \\
        y_f \\
        z_f \\
        1
    \end{pmatrix}_{car} = \Ek^{-1} \times \begin{pmatrix}
        x_f \\
        y_f \\
        z_f \\
        1
    \end{pmatrix}_k
\end{equation}

Finally, we create the bird's eye view representation $\mathbf{B} \in \mathbb{R}^{N_f \times X \times Y}$ from this 3D point cloud containing the image features by following the voxel sum pooling method proposed in LSS \cite{philion-lss}. This results in the 2D BEV map that will be later used for the prediction of semantic grids. 

\begin{table*}[t]
\centering

\begin{tabular}{@{}l|c|ccccc|c@{}}
\toprule
Method                        & Modalities & Human          & Vehicle        & Movable Object   & Drivable Area  & Walkway        & FPS   \\ \midrule
LAPTNet                       & C+L        & 13.8           & 40.13          & 27.4           & 79.43          & 57.25          & \textbf{43.8} \\
LAPTNet-FPN                   & C+L        & 22.17 & 48.04 & 32.2  & 81.78 & 61.25 & \textbf{\color[HTML]{3531FF}38.1}  \\ 
LAPT-PP                       & C+L        & \textbf{\color[HTML]{3531FF}30.45}          & 50.15          & \textbf{\color[HTML]{3531FF}36.31}          & 81.88          & 60.26          & 32.2  \\
LAPT-FPN-PP                   & C+L        & \textbf{33.92} & \textbf{\color[HTML]{3531FF}52.4}  & \textbf{38.38} & \textbf{\color[HTML]{3531FF}83.04} & \textbf{\color[HTML]{3531FF}62.11} & 25.1  \\ \midrule
Pyramid Occupancy Network \cite{roddick-pon} & C          & 8.2            & 15.37          & 6.9            & 60.4           & 31.0           & 22.3  \\
Lift-Splat-Shoot  \cite{philion-lss} & C          & 9.99           & 32.02          & 21.6           & 72.9           & 51.03          & 28.3  \\
M2BEV  \cite{m2bev} & C          & \textit{-}     & -              & \textit{-}     & 75.9           & -              & 4.3   \\
FIERY   \cite{fiery}       & C+T        & \textit{-}     & 35.8           & \textit{-}     & \textit{-}     & \textit{-}     & -     \\
Translating Images into Maps \cite{imgs2map}& C          & 8.7            & 38.9           & 13.2           & 72.6           & 32.4           & -     \\
BEVFormer (Seg) \cite{bevformer}   & C          & -              & 44.8           & -              & 77.5           & -              & -     \\
Harley et al.    \cite{harley-bev} & C        & -              & 47.0          & -              & -              & -              & 7.3      \\\midrule
PointPillars baseline                  & L        & 0.0            & 24.99          & 16.81          & 58.44          & 33.66          & -     \\
FISHINGNet     \cite{hendy2020fishing} & C+L        & 20.4           & 40.9           & -              & -              & -              &  -    \\
TFGrid     \cite{transfusegrid}   & C+L        & -              & 35.88          & -              & 78.87          & 50.98          & 18.3 \\ 
Harley et al.    \cite{harley-bev} & C+L        & -              & \textbf{60.8 }          & -              & -              & -              & -      \\ 
BEVFusion    \cite{bevfusion} & C+L        & -              & -          & -              & \textbf{85.5}              & \textbf{67.6}              & -      \\ \bottomrule

\end{tabular}

\caption{IoU [\%] on the validation split for the nuScenes dataset. Best results are in bold, second best are in blue font.}
\vspace{-8mm}
\end{table*}

\subsection{Multi-scale LAPT}
\label{subsec:laptfpn}

In the proposed LAPT method, we specify that the projection is only performed with the deepest feature map available from the network. But, in practice, using only this feature map yields a very sparse BEV representation. Since convolutional backbones gradually reduce the spatial size of the original image, we follow the idea proposed in FPN \cite{fpn} and project different image scales in order to generate a denser BEV map. For LAPT-FPN, we project the two last scales of the feature maps delivered by the chosen backbone to a BEV of spatial size $(X \times Y)$. 

Having multiple scales to project to the 3D space means that the resulting sparse depth image $\Dk$ needs to be downsampled using the corresponding kernel values $d_f$ for each scale. It also implies that the steps described in Eq. \ref{eq:transform2Dto3D}-\ref{eq:transform_TFs} need to be performed for each of the chosen scales to generate separate $\mathbf{B} \in \mathbb{R}^{C \times X \times Y}$ maps which need to be fused together to generate a unified BEV representation. For the case of LAPT-FPN, we sum them together to generate the unified BEV representation containing all the image features.

\subsection{BEV decoding}
\label{subsec:bevencode}

Given that the BEV representation has an image-like structure, we use also a convolutional network to perform the decoding. Specifically, we use a residual convolutional network \cite{resnet} together with an upsampling block made of a bilinear upsampling step, ($3\times3$) convolutions and ReLU activation to have the desired grid spatial size of $(X,Y)$. One final $(1\times1)$ convolution is applied to predict the desired semantic grid $\mathbf{y} \in \mathbb{R}^{C \times X \times Y}$.

\section{EXPERIMENTAL SETUP}

Here we discuss the experimental setup to train our approach, as well as the proposed experiments that explore the effectiveness of the network for semantic grid prediction. Following previously mentioned sensor-fusion approaches \cite{bevfusion,transfusegrid}, we also explore how the network can be improved by adding a parallel backbone to generate a BEV map using only LiDAR information as well as different ways of fusing the image-based and LiDAR-based top-view representations for semantic grid prediction. 

\subsection{Dataset used}

We use the nuScenes dataset for our experiments \cite{nuscenes}. nuScenes is a large-scale dataset for different tasks related to autonomous driving. It is made of 1000 driving scenes taken in the cities of Boston and Singapore, where each scene has a duration of 20 seconds with synchronized keyframes containing LiDAR, camera and radar data at a sampling rate of 2Hz. Their setup of six cameras enables a 360$^{\circ}$ surround view of each scene. Aside from the sensor data, this dataset offers 3D bounding box annotations as well as HD maps.

To generate the ground truth for training, we take the information available from the HD maps and 3D bounding boxes and project them into a BEV representation with the vehicle in the center. The grid space represents a square of 100m by 100m at a 0.5m resolution with the car in its center, resulting in ground truth samples with a $(X,Y)$ shape of $(200\times200)$ pixels. We choose 5 classes from these annotations: drivable area, walkway, human, vehicle and movable object to facilitate comparisons to the state-of-the-art. The ground truth labels for the drivable area and walkway classes are rasterized from the HD maps. The human, vehicle and movable object ground truths are generated by projecting their corresponding cuboid annotations to the BEV.


\subsection{Training setup and evaluation metrics}

We train our network to predict each class separately $(C=1)$. We start with an ImageNet \cite{deng2009imagenet} pretrained EfficientNet-B0 and we perform training on both the camera encoder and BEV decoder. Before backbone inference, we concatenate images along the batch dimension, resulting in an input tensor of size $(B*C, 3,H,W)$. As loss function, we follow LSS \cite{philion-lss} and use the binary cross-entropy loss. The Adam \cite{adam-optim} optimizer is used with a learning rate of $1e-3$, weight decay of $1e-7$ and the standard values for the betas and epsilon given in the Pytorch implementation. We train all of our networks with batch size of 16 samples for 100 epochs or until intersection over union (IoU) convergence in the validation split is reached, whichever happens first. Each sample contains the 6 images from the surround camera setup and the point cloud from the LiDAR for a given keyframe. For evaluation, we use the commonplace metric of intersection over union for segmentation. We use EfficientNet-B0 as our image encoder and a ResNet18 as our BEV decoder together with the mentioned upsampling blocks. We only use one LiDAR sweep and resize input images to a resolution of $(128\times352)$ pixels.

\subsection{Model modifications, ablation and adverse conditions studies}

We evaluate the difference between using only the last scale ($d_f=16$) of the camera encoding network, and using the last two scales ($d_f \in \{8,16\}$), we name these models LAPT and LAPT-FPN respectively. We also evaluate how the performance of the model changes if instead of projecting the smallest feature map, with ($d_f=16$), directly to the final BEV space, we project it to a grid of half the final resolution ($\mathbf{B} \in \mathbb{R}^{N_f \times X/2 \times Y/2}$) which is then upsampled by $2\times$ using the same upsampling architecture as in the BEV decoder to match the final grid size. We denote the models that project to a coarser grid with (MS$_{\text{B}}$).

We train and evaluate a point-cloud-only baseline method with PointPillars as the encoder and our proposed BEV decoder, in order to compare how much the use of camera features influences the quality of the predicted grids. Since our method uses point clouds, we also evaluate how the quality of the final grid is affected if a parallel backbone is added to process them separately. We use the PointPillars \cite{pointpillars} backbone to generate a BEV representation from the LiDAR input, as in \cite{pillarsegnet} and \cite{transfusegrid}, to fuse it through addition with our unified BEV representation with the camera features. We name these methods LAPT-PP and LAPT-FPN-PP depending on the amount of feature map scales used. Motivated by \cite{hendy2020fishing}, we evaluate as well three different methods of fusing together the BEV representations generated from each modality: sum, concatenation along the channel axis and ($1\times1$) maxpooling along the channel dimension of the two concatenated feature maps. 

Finally, we explore how our methods fare in adverse conditions such as night and rain, which are available in the nuScenes dataset. These two conditions were chosen to further evaluate the dependence to each sensor modality, where the night condition should challenge the camera modality \cite{cv-book} and the rain condition should challenge the LiDAR modality \cite{lidar-book}.

\section{RESULTS AND DISCUSSION}

\subsection{Comparison with the state of the art}

In table I, a comparison to other state of the art methods can be seen using the IoU metric. Here we can observe that by guiding the projection of the features using real geometric information about the scene, our multiscale approach (LAPT-FPN) network is able to outperform all other networks which need to learn the relationships between the camera plane and the BEV. Even comparing against methods that perform sensor fusion such as FISHINGNet \cite{hendy2020fishing} and TransFuseGrid \cite{transfusegrid}, LAPT-FPN is able to outperform them across the available classes without having to add a LiDAR-specific backbone to process point clouds. 

For LAPT-FPN, we report improvements over the state of the art of 8.67\% for human segmentation \cite{hendy2020fishing} and 49.07\% for movable object segmentation \cite{philion-lss}. We achieve similar results to \cite{bevfusion} for drivable area (-4\%) and walkway (-10.3\%) segmentation for inference on images with 1/4 the resolution. We achieve a lower (-26.5\%), segmentation value for the vehicle class compared to \cite{harley-bev} which uses an image resolution $9.5\times$ bigger than ours and reports a latency $5.2\times$ slower for the RGB-only method. We theorise that the difference of performance with these approaches might be related to the use of deeper networks for \cite{harley-bev} or transformers for \cite{bevfusion}, to extract features from the images, as well as the use of bigger images.
\vspace{-0.3mm}

\subsection{Addition of LiDAR-specific encoder}
\label{sub:res-addition-lidar}
With the addition of a LiDAR-specific encoder based on PointPillars, the performance of our proposed methods is improved even further. With our proposed architecture LAPT-FPN-PP, we improve the state of the art by 66.27\% for human segmentation and 77.68\% for movable object segmentation. We achieve a close second place against \cite{bevfusion} for drivable area (-2.4\%) and walkway (-8.8\%) segmentation. Using this encoder we narrow the gap with \cite{harley-bev} in vehicle segmentation (-16\%) while being $3.7\times$ faster.
\vspace{-0.3mm}

\subsection{Using multiple scales and MS$_{\text{B}}$ projection}
\begin{figure}[h]
  \centering
  \includegraphics[width=\linewidth]{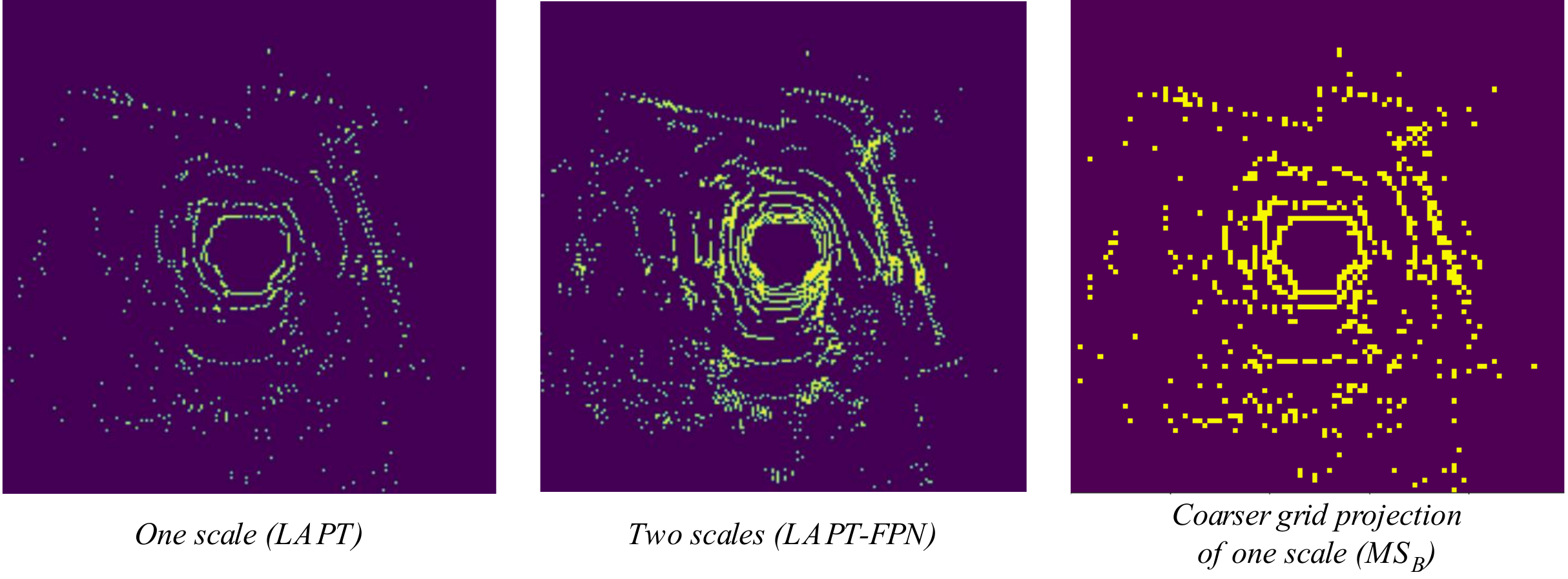}
  \caption{Visualization of the number of points projected to the BEV when using one (left) or two (center) image scales. We also visualize the idea of projection to a grid half the resolution of the base approach (left). Best viewed with digital zoom.}
  \vspace{-5mm}
  \label{fig:scales-LAPT}
\end{figure}

In Fig. \ref{fig:scales-LAPT}, we show the amount of points that are projected for each feature scale in a random sample. We also show how the projection approach to a coarser grid (MS$_{\text{B}}$) compares to the original approaches.

We can see in Tab. II, how the performance of the grid is affected by projecting the features with ($d_f=16$) to a grid with half the resolution of the original one proposed in LAPTNet. Here, we can see that this extra projection and upsampling step does not bring any improvement over the LAPT and LAPT-FPN models for the evaluated classes. The proposed MS$_{\text{B}}$ step only brings marginal improvements for the LAPT-PP and LAPT-FPN-PP methods.

\setlength{\tabcolsep}{3pt}

\begin{table}[h]
\centering
\begin{tabular}{@{}lccccc@{}}
\toprule
                 & \multicolumn{1}{l}{Human} & \multicolumn{1}{l}{Vehicle} & \multicolumn{1}{l}{Mov. Obj.} & \multicolumn{1}{l}{Driv. A} & \multicolumn{1}{l}{Walkway} \\ \midrule
LAPT             & 13.8                    & 40.13                     & 27.45                          & 79.43                           & 57.25                     \\
LAPT (MS$_{\text{B}}$)        & 3.42                    & 30.61                     & 0.0                            & 67.09                           & 40.04                     \\
LAPT-FPN         & \textbf{22.17}          & \textbf{48.04}            & \textbf{32.2}                  & \textbf{81.78}                  & \textbf{61.25}            \\
LAPT-FPN (MS$_{\text{B}}$)    & 8.57                    & 42.07                     & 9.05                           & 78.03                           & 52.52                     \\ \midrule
LAPT-PP          & 30.11                   & 50.14                     & 35.86                          & 81.89                           & 60.25                     \\
LAPT-PP (MS$_{\text{B}}$)     & 33.0                    & 52.4                      & 38.11                          & 83.03                           & 62.11                     \\
LAPT-FPN-PP      & \textbf{33.25}          & \textbf{52.4}             & \textbf{38.11}                 & 83.03                           & 62.11                     \\
LAPT-FPN-PP (MS$_{\text{B}}$) & 32.74                   & 51.75                     & 37.06                          & \textbf{83.07}                  & \textbf{62.52}            \\ \bottomrule 
\end{tabular}
\caption{IoU [\%] comparison when using BEV multiscale (MS$_{\text{B}}$).}
\vspace{-8mm}
\end{table}

\subsection{Fusion methods for BEV maps from LiDAR and camera}

As previously stated, following the study presented in \cite{hendy2020fishing}, we study how different fusion methods can impact the performance of the network when dealing with separate BEV representations that encode camera and LiDAR features. We present the results for each of the proposed fusion methods (sum, concatenation and maxpooling) in both LAPT-PP and LAPT-FPN-PP in table III. For LAPT-PP, we find that only the maxpooling-based fusion improves the detection of humans by 12.02\%, and for the rest of classes the performance does not change greatly with the other fusion methods. In the case of LAPT-FPN-PP, any of the other two fusion approaches (concat and maxpool), do to not bring any improvements to the creation of semantic grids.

\begin{table}[h]
\centering
\begin{tabular}{lccccc}
\toprule
           & \multicolumn{1}{l}{Human} & \multicolumn{1}{l}{Vehicle} & \multicolumn{1}{l}{Mov Obj.} & \multicolumn{1}{l}{Driv. A} & \multicolumn{1}{l}{Walkway} \\ \midrule
LAPT-PP sum        & 30.11                   & \textbf{50.14}            & 35.86                      & 81.89                & \textbf{60.25}            \\
LAPT-PP concat     & 30.19                   & 49.72                     & 36.08                      & 79.39                & 56.56                     \\
LAPT-PP maxpool    & \textbf{33.73}          & 49.36                     & 35.92                      & 78.34                & 56.33                     \\ \midrule
LAPT-FPN-PP sum        &  \textbf{33.25}          &  52.4                      &  \textbf{38.11}             &  83.03                &  62.11                     \\
LAPT-FPN-PP concat     &  33.64                   &  \textbf{52.65}            &  37.91                      &  81.09                &  58.66                     \\
LAPT-FPN-PP maxpool    &  29.82                   &  52.58                     &  37.58                      &  78.95                &  61.34                     \\ \bottomrule 
\end{tabular}
\caption{IoU [\%] for LAPT-PP and LAPT-FPN-PP ablation study using the three proposed fusion methods}
\vspace{-6mm}
\end{table}

\subsection{Robustness against adverse conditions}

In Tab. IV, we report the performance of our network on the scenes under rain and night conditions for our proposed methods. As a general trend, we see that the network is affected more by the night condition than rain. For us, this suggests that the network relies more on camera information, such as color or texture, that is difficult to capture at night, than the geometric information from the point cloud, which might be noisy when it is raining. As noted in subsection \ref{sub:res-addition-lidar} we also observe that the addition of a LiDAR encoder helps to mitigate the effects in both types of adverse conditions.

Comparing to the general results in the validation set we can see that even for the best performing method (LAPT-FPN-PP), the performance drops sharply on night scenes. In this case, the network seems to perform worse with classes which have a smaller grid footprint, like human and movable object, dropping their IoU scores by two thirds. For the walkway class, performance also drops significantly during the night (-46\%). However, this does not seem to be the case for vehicles under any of the two adverse circumstances. The system seems to be robust against rainy conditions, achieving segmentation results similar to the ones in the general validation split across all methods and classes. It is important to note here that there is a difference in scores with adverse conditions versus the whole validation split since we are evaluating a robust model on a smaller subset of the data.



\begin{table}[h]
\centering
\begin{tabular}{@{}lccccc@{}}
\toprule
                    & \multicolumn{1}{l}{Human} & \multicolumn{1}{l}{Vehicle} & \multicolumn{1}{l}{Mov Obj.} & \multicolumn{1}{l}{Driv. A} & \multicolumn{1}{l}{Walkway} \\ \midrule
LAPT (Rain)         & 10.14          & 44.74          & 32.39          & 73.08          & 49.96          \\
LAPT-FPN (Rain)     & 17.40          & 51.54          & 36.92          & 74.48          & 52.50          \\
LAPT-PP (Rain)      & 22.14          & 52.79          & 37.58          & 75.20          & 51.66          \\
LAPT-FPN-PP (Rain)  & 23.58          & 55.13          & 40.78          & 77.12          & 53.33 \\ \midrule
LAPT (Night)        & 6.48           & 36.79          & 12.23          & 67.89          & 25.93          \\
LAPT-FPN (Night)    & 11.67          & 46.52          & 13.87          & 71.02          & 28.76          \\
LAPT-PP (Night)     & 10.80          & 50.21          & 15.06          & 73.39          & 34.56          \\
LAPT-FPN-PP (Night) & 13.79          & 52.45          & 12.06          & 74.55          & 33.16     \\ \midrule
LAPT-FPN-PP (All)    &  33.92         & 52.4           & 38.38          & 83.04          & 62.11          \\ \bottomrule
\end{tabular}
\caption{IoU [\%] for the adverse conditions study}
\vspace{-8mm}
\end{table}

\subsection{Semantic grid generation latency}

An autonomous system should aim to have as little latency as possible in order to react accordingly to changes in its environment \cite{robotics-book}. This implies that the processing time for each part of the autonomy stack is crucial. Motivated by this, we compare the latency of our proposed methods to some of the ones currently available in the literature.

The inference time in frames per second (FPS) is presented for the different LAPTNet variations as well as other methods of the state of the art in Tab. I. These results are reported on a Nvidia V100 GPU. Comparing against LSS, our best performing method achieves a similar latency with much better segmentation results. Comparing against some state of the art approaches based on attention mechanisms \cite{harley-bev}, we see that using the geometric information available from the LiDAR sensor results in a $6\times$ speedup for the case of our fastest method (LAPT) or a $3.7\times$ speedup for our best perfoming method (LAPT-FPN-PP). 

Looking at the sensor setup used to capture the nuScenes dataset, we see that any of our proposed approaches would be able to keep up with the slowest sensor in the stack, making our system run in real time. Here, the LiDAR sensor would be the slowest sensor in the vehicle running at 20Hz \cite{nuscenes}.


\section{CONCLUSIONS}

We have presented in this paper the multiscale LiDAR-aided Perspective Transform Network (LAPT-FPN). By using information available in point clouds, the LAPT-FPN is able to correlate image features extracted by a convolutional backbone to a bird's eye view representation for semantic grid prediction. Doing this correlation across different image scales provided by the backbone, allows the method to outperform some state of the art approaches while being able to run in real time.  
\vspace{-0.3mm}
\section*{ACKNOWLEDGMENT}
\small
The experiments presented in this paper were carried out using the Grid'5000 testbed, supported by a scientific interest group hosted by Inria and including CNRS, RENATER and several Universities as well as other organizations.

\bibliographystyle{IEEEtran}
\bibliography{egbib}   

\end{document}